\documentclass{llncs}
\usepackage[T1]{fontenc}
\usepackage{graphicx}
\usepackage{todonotes}
\usepackage[colorlinks=true, linkcolor=blue, citecolor=blue, urlcolor=blue]{hyperref}
\usepackage{booktabs}
\usepackage{array}
\usepackage{placeins}
\usepackage{flafter}

\begin{document}

\title{Region-Guided Food-Package Evidence Extraction and Abstention-Aware Draft R3337 Screening with Selective VLM Review}

\titlerunning{Region-Guided R3337 Screening with Selective VLM Review}
%
%

\author{
Mayimunah Nagayi\inst{1}\textsuperscript{*}\orcidID{0009-0009-9241-3059}
\and
Alice Scaria Khan\inst{2}\orcidID{0000-0003-1011-3693}
\and
Tamryn Frank\inst{2}\orcidID{0000-0002-5180-9171}
\and
Rina Swart\inst{3}\orcidID{0000-0002-7786-3117}
\and
Clement Nyirenda\inst{1,4}\orcidID{0000-0002-4181-0478}
}

\authorrunning{M. Nagayi et al.}

\institute{
Department of Computer Science, University of the Western Cape,
Cape Town, South Africa\\
\email{4163113@myuwc.ac.za}\textsuperscript{*}
\and
School of Public Health, University of the Western Cape,
Cape Town, South Africa\\
\email{\{askhan,tfrank\}@uwc.ac.za}
\and
Department of Dietetics and Nutrition, University of the Western Cape,
Cape Town, South Africa\\
\email{rswart@uwc.ac.za}
\and
eResearch Office, University of the Western Cape,
Cape Town, South Africa\\
\email{cnyirenda@uwc.ac.za}
}

\maketitle              
%

\begin{abstract}
Applying South Africa’s draft R3337 warning criteria to food-package photographs requires nutrition, ingredient and product-form evidence. This study presents a region-guided evidence extraction and abstention-aware screening workflow for draft R3337. A three-class detector localises nutrition information panels, ingredient lists and product-identity regions, after which PaddleOCR and structured extraction recover nutrient values, ingredient evidence and product-form information. Recovered evidence is aggregated and assessed using deterministic draft R3337 rules. When evidence is incomplete, uncertain or conflicting, the workflow abstains through \textit{Review} or \textit{Insufficient evidence} rather than forcing an automatic outcome. For qualifying unresolved cases, a selective vision-language model (VLM) review stage using Qwen3-VL 8B Instruct reassesses unresolved evidence. Accepted evidence is merged and the same deterministic rules are reapplied. The workflow was applied to 442 products represented by 3,929 images. Evaluation used a manual validation set of 65 products. Before selective VLM review, 31 of 60 decidable products received automatic outcomes and 30 were correct, giving 51.7\% coverage and 96.8\% accuracy among automatic outcomes (selective accuracy). Selective VLM review resolved eight additional products, increasing coverage to 65.0\%. Of 39 automatic outcomes, 38 were correct, giving 97.4\% selective accuracy. The single false warning was retained from the rules-only workflow, while no unsafe warning omissions occurred after selective review. Evidence-level evaluation nevertheless identified incorrect VLM-recovered nutrition, ingredient and product-form evidence. These findings show that selective VLM review can recover unresolved evidence, while deterministic screening and explicit abstention remain necessary for reliable image-based draft R3337 screening.


\end{abstract}

\section{Introduction}

Front-of-pack labelling (FOPL) is intended to make important nutrition information more visible to consumers at the point of purchase. The World Health Organization~\cite{who2019fopl} identifies FOPL as a policy tool that can support healthier food choices. Bopape et al.~\cite{bopape2021perceptions} found that South African consumers viewed warning labels positively and described them as concise and understandable. The Department of Health~\cite{doh2023r3337} published the Draft Regulations Relating to the Labelling and Advertising of Foodstuffs (R3337), which propose front-of-pack warning symbols for prepackaged foods that meet specified conditions relating to total sugar, saturated fat, sodium and artificial sweeteners. Applying these criteria requires nutrient values, evidence of added nutrient sources, product form and, where applicable, nutrition information after preparation.

Previous South African draft R3337 assessments used captured or manually coded product information~\cite{abdoolkarim2025compliance,mamabolo2026warning}, whereas direct screening from food-package photographs first requires the relevant evidence to be recovered from the images. The information required for screening may appear across several photographs and package surfaces, while small text, dense layouts, glare, curved packaging and other image conditions can affect text recognition. Recognising package text alone is also insufficient. Nutrient names must be associated with the correct values, units and reference bases, while ingredient and product-form evidence must be recovered and combined at product level. Image-derived evidence may be incomplete, uncertain or conflicting, making it unsafe to force an automatic screening outcome when the required evidence cannot be established reliably.

This study develops an integrated region-guided, abstention-aware workflow for screening food-package images against draft R3337. It combines region localisation, optical character recognition (OCR), structured evidence extraction, deterministic screening and selective vision-language model (VLM) review, while retaining \textit{Review} or \textit{Insufficient evidence} when evidence remains unresolved. The main contributions of this study are as follows:

\begin{enumerate}
    \item A region-guided workflow for recovering and integrating nutrition, ingredient and product-form evidence from real South African food-package photographs.
    
    \item A deterministic implementation of the draft R3337 warning criteria with explicit abstention when image-derived evidence is incomplete, uncertain or conflicting.
    
    \item A selective VLM evidence-review stage that reassesses qualifying unresolved evidence before deterministic rule reapplication.
    
    \item A product-level evaluation of the proposed screening workflow against manual reference outcomes, together with an evidence-level evaluation of the reliability of VLM-recovered nutrition, ingredient and product-form evidence.
\end{enumerate}

The remainder of this paper is organised as follows. Section~2 reviews related work and identifies the research gap. Section~3 describes the datasets. Section~4 presents the methodology. Section~5 presents the results and evaluation. Section~6 discusses the findings, Section~7 presents the study limitations and Section~8 concludes the paper.

\section{Related Work}

\subsection{Front-of-Pack Warning Labels and Draft R3337 Assessment}

Nutrient profiling models classify packaged foods according to their nutritional composition and nutrients of concern. Frank et al.~\cite{frank2022nutrient} applied four nutrient profiling models to the South African packaged-food supply and found that classifications differed across the models. Draft R3337 specifies warning criteria based on total sugar, saturated fat, sodium and artificial sweeteners, together with conditions relating to added nutrient sources, product form and, where applicable, prepared-product values~\cite{doh2023r3337}. Abdool Karim et al.~\cite{abdoolkarim2025compliance} used coded package and nutrition variables from 6,768 photographed products to determine warning-label eligibility and assess on-package marketing under draft R3337. Mamabolo et al.~\cite{mamabolo2026warning} assessed 271 breakfast porridges and cereals and found that warning-label requirements varied with preparation method and grain type. These studies show the importance of nutrient, ingredient and preparation-related evidence, but did not automate its recovery from food-package photographs.

\subsection{Food-Package Evidence Recovery}

Food-package photographs present challenges that differ from scanned documents, including curved and reflective surfaces, visually crowded layouts and uncontrolled image conditions. Guimarães et al.~\cite{guimaraes2023review} reviewed grocery-label detection and recognition and identified product variation, limited datasets and uncontrolled image conditions. Nagayi et al.~\cite{nagayi2026ocr} evaluated four OCR systems on real South African food-package images and identified dense layouts, multilingual text, glare, curved surfaces and small text as packaging-specific challenges. Their evaluation focused on text recovery, while Nagayi et al.~\cite{nagayi2026r214} combined region detection, OCR-derived sodium evidence, product identification and deterministic R214 screening.

Region localisation can reduce irrelevant package content before text recognition. Saputra et al.~\cite{saputra2024bounding} used a MobileNetV1-based detector to locate nutrition-label regions before applying preprocessing and Tesseract OCR. Their evaluation was limited to ten images and focused on extracting text from nutrition labels. Fuentes et al.~\cite{fuentes2022transcription} developed an OCR-based system that preserved spatial relationships between nutrient names and quantities when extracting nutritional information from Spanish food products. Their approach demonstrates the importance of retaining layout information when associating nutrient labels with the correct values. Pettersson et al.~\cite{pettersson2024multimodal} combined visual features with OCR-derived text for fine-grained grocery-product recognition and found that the two information sources complemented each other. Food-package evidence recovery therefore involves more than isolated text recognition and can depend on region localisation, spatial association and integration of visual and textual information.

\subsection{Vision-Language Models, Reliability and Selective Prediction}

Vision-language models (VLMs) can perform text recognition, visual question answering and structured information extraction from image and text inputs. Liu et al.~\cite{liu2024ocrbench} evaluated multimodal models using OCRBench across text-related image tasks and reported useful capabilities alongside continuing weaknesses in multilingual and other challenging text settings. Assiri et al.~\cite{assiri2025nutrition} evaluated multimodal language models for extracting nutritional elements and values from bilingual food-label images, demonstrating their relevance to food-package information extraction. The Qwen3-VL family described by Bai et al.~\cite{bai2025qwen3vl} supports combined image and text inputs and includes an 8-billion-parameter variant. However, performance on general benchmarks or related extraction tasks does not establish reliable recovery of exact nutrient values, units, reference bases and ingredient evidence from real food-package images. Selective prediction provides an approach for handling uncertain model outputs. Geifman and El-Yaniv~\cite{geifman2019selectivenet} introduced SelectiveNet, which jointly considers prediction and rejection while examining coverage and error risk. Whitehead et al.~\cite{whitehead2022reliable} applied abstention to visual question answering, where withholding an answer was preferred when confidence was insufficient. These studies show the value of allowing uncertain cases to remain unresolved rather than requiring a prediction for every input. In evidence-extraction settings, however, abstention does not establish that the evidence returned is factually correct. The reliability of recovered evidence therefore remains important alongside coverage and abstention.

\subsection{Research Gap}

Previous work has addressed draft R3337 assessment, food-package OCR and region localisation, structured nutrition extraction, multimodal food-label extraction and selective prediction. Nagayi et al.~\cite{nagayi2026r214} integrated image-derived evidence for R214 sodium screening, but not the broader nutrition, ingredient and product-form evidence required for draft R3337. The reviewed studies do not evaluate draft R3337 screening with explicit abstention and selective VLM review. The contribution is therefore the system-level integration and evaluation of these components rather than the development of a new detector or VLM architecture.

\section{Dataset Description}

\subsection{Food-Package Screening Dataset}

The study uses real-world food-package photographs from the South African Nutrition Facts Panel project of the Department of Dietetics and Nutrition at the University of the Western Cape (UWC). The photographs were collected in 2023 by trained personnel. The screening dataset contains 442 packaged-food products represented by 3,929 full-package photographs. Each product is represented by between three and twenty-seven photographs, averaging 8.9 photographs per product. The photographs capture different package surfaces, meaning that product identity, nutrition information, ingredient information and preparation details may appear in different images of the same product. The product is the unit of analysis. All photographs are linked to their corresponding product and are treated as sources of information rather than as independent samples. The screening dataset is separate from the annotated dataset used to develop and evaluate the region detector described in Section~3.2.

\subsection{Detector-Development Dataset}

For detector development, 664 annotated food-package images from 176 products were used, containing 885 bounding boxes across three classes, namely nutrition information panels, ingredient lists and product-identity regions. The annotations comprised 201 nutrition-panel regions, 274 ingredient-list regions and 410 product-identity regions. The dataset was divided at product level into training, validation and held-out test subsets with an approximately 80\%, 10\% and 10\% split. The training subset contained 141 products (80.1\%), 524 images and 700 bounding boxes. The validation subset contained 18 products (10.2\%), 89 images and 120 bounding boxes, while the held-out test subset contained 17 products (9.7\%), 51 images and 65 bounding boxes. No product appeared in more than one subset, ensuring that evaluated products were not represented during training.

\section{Methodology}

\subsection{Overall Workflow}

\begin{figure}[!htbp]
    \centering
    \includegraphics[width=\linewidth]{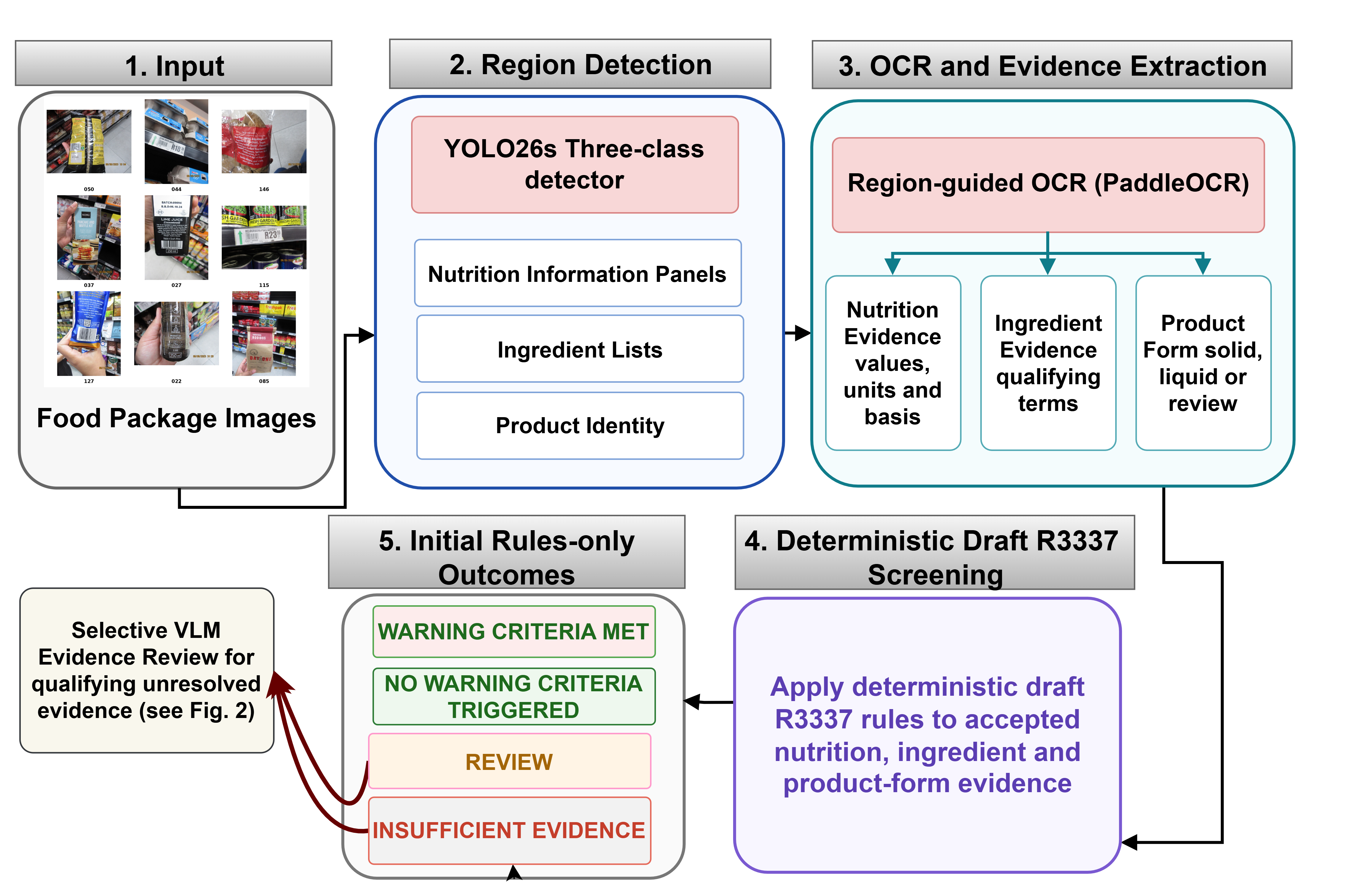}
    \caption{Initial region-guided evidence extraction and deterministic draft R3337 screening workflow.}
    \label{fig:primary_workflow}
\end{figure}

The proposed methodology consists of an initial region-guided screening workflow followed by selective VLM evidence review for qualifying unresolved evidence. Figure~\ref{fig:primary_workflow} presents the initial stages from the food-package images described in Section~3 through region detection, optical character recognition (OCR), evidence extraction and deterministic draft R3337 screening.

\begin{enumerate}
    \item \textbf{Input.} The workflow receives the full-package photographs associated with each product.

    \item \textbf{Region detection.} A three-class detector localises nutrition information panels, ingredient lists and product-identity regions.

    \item \textbf{OCR and evidence extraction.} Region-guided OCR recovers nutrition, ingredient and product-form evidence, while unresolved evidence is retained.

    \item \textbf{Deterministic draft R3337 screening.} Product-level evidence is assessed using the encoded rules without forcing outcomes when required evidence remains uncertain or unavailable.

    \item \textbf{Initial rules-only outcomes.} Products are assigned \textit{Warning criteria met}, \textit{No warning criteria triggered}, \textit{Review} or \textit{Insufficient evidence}, with the latter two representing abstention.
\end{enumerate}

Qualifying unresolved evidence proceeds to the selective VLM evidence-review stage described in Section~4.5 and illustrated in Figure~\ref{fig:vlm_review_workflow}. Accepted evidence is merged before the same deterministic draft R3337 rules are reapplied.

\subsection{Region Detection and Optical Character Recognition}

The annotated detector-development dataset described in Section~3.2 is used to train a three-class You Only Look Once version 26 small (YOLO26s) object detector from pretrained weights using Ultralytics version 8.4.89~\cite{jocher2026yolo26}. Training uses automatic optimiser selection, deterministic execution with a seed of 0 and automatic mixed-precision computation. The checkpoint selected at epoch 77 based on validation performance is retained for evaluation on the held-out test subset and inference on the food-package screening dataset. During inference, images are processed at an input size of \(640 \times 640\) pixels using a confidence threshold of 0.25. Each retained detection preserves the product and source-image identifiers, region class, bounding-box coordinates and detector confidence.

OCR is applied to the detected nutrition-panel, ingredient-list and product-identity regions using PaddleOCR version 3.5.0 with the graphics processing unit (GPU) build of PaddlePaddle version 3.3.0~\cite{cui2025paddleocr}. The configuration uses the \texttt{PP-OCRv5\_server\_det} text-detection model, the \texttt{en\_PP-OCRv5\_mobile\_rec} English recognition model and the \texttt{PP-LCNet\_x1\_0\_textline\_ori} text-line-orientation model, with text-line orientation enabled. Recognised text from ingredient-list and product-identity regions is retained with its product and source-image information, while nutrition-panel outputs retain recognised tokens and their coordinates for structured extraction. Each nutrition-panel region is processed in its original form and in a second rendering produced by adding approximately 10\% padding and enlarging the padded region by a factor of two. Source photographs, detected crops, alternative renderings and OCR outputs remain linked to the originating product and are not treated as independent product samples.

\subsection{Structured Evidence Extraction and Product-Level Aggregation}

\textit{Nutrition evidence.} Recognised nutrition-panel tokens and their coordinates are processed using a header-aware, coordinate-based extractor that associates nutrient labels with numerical values, units and reference bases. Candidate values are identified using nutrient-label matching, row proximity and table-header interpretation. The warning-related targets are total sugar, saturated fat and sodium, while total fat is retained as supporting information rather than assessed as a draft R3337 warning component. Consistency checks across the original and padded renderings consider nutrient identity, numerical value, unit and reference basis. Evidence remains unresolved when the row association, value, unit or reference basis cannot be established reliably, or when renderings or nutrition panels produce conflicts. Salt values are retained separately and are not compared directly with sodium thresholds. When an explicit sodium value is unavailable, salt-derived sodium evidence is accepted only when the conversion, unit and applicable per-100 reference basis are clear. Preparation information is considered where available, but evidence remains unresolved when the applicable basis cannot be established reliably.

\textit{Ingredient evidence.} Recognised text from ingredient-list regions is grouped by product and searched using deterministic term dictionaries and text-matching rules. The categories cover added sugar, added sodium, added saturated-fat sources and artificial sweeteners. Added saturated-fat evidence is restricted to qualifying dictionary terms and does not include every ingredient that may naturally contain saturated fat. Each category is assigned \textit{Yes}, \textit{Not detected} or \textit{Uncertain}. \textit{Yes} indicates that a configured qualifying term is identified in recognised text. \textit{Not detected} indicates that no configured term is found and is not interpreted as proof that the ingredient is absent. \textit{Uncertain} is assigned when recognised text or the match cannot support a clear result. Missing ingredient-list regions and empty OCR outputs remain unavailable evidence.

\textit{Product form and product-level aggregation.} Product form is determined from product-identity information and package text, as the applicable draft R3337 nutrient thresholds differ for solid and liquid products. The product-form assessment returns \textit{Solid}, \textit{Liquid} or \textit{Review}, with \textit{Review} retained when the available information is insufficient or conflicting. Recovered nutrition, ingredient and product-form evidence is aggregated at product level. Where evidence remains missing, uncertain or conflicting across available sources, that uncertainty is preserved rather than resolved by forcing a value or classification. The product-level evidence is passed to deterministic draft R3337 screening in Section~4.4.

\subsection{Deterministic Draft R3337 Screening and Abstention}

The product-level evidence produced in Section~4.3 is assessed using deterministic rules that operationalise the draft R3337 warning criteria. For consistent implementation, the study applies nutrient-specific conjunctive rules to total sugar, saturated fat and sodium. A nutrient-related warning requires both qualifying added-source evidence for the corresponding nutrient and a nutrient value greater than the applicable solid or liquid threshold. The implementation uses strict greater-than comparisons consistent with the draft R3337 nutrient cut-offs~\cite{doh2023r3337}.

For solid products, the encoded thresholds are:
\begin{itemize}
    \item total sugar greater than 10.0~g per 100~g.
    \item saturated fat greater than 4.0~g per 100~g.
    \item sodium greater than 400~mg per 100~g.
\end{itemize}

For liquid products, the encoded thresholds are:
\begin{itemize}
    \item total sugar greater than 5.0~g per 100~ml.
    \item saturated fat greater than 3.0~g per 100~ml.
    \item sodium greater than 100~mg per 100~ml.
\end{itemize}

The artificial-sweetener criterion requires explicit evidence of an added artificial sweetener and does not require a corresponding nutrient-table value~\cite{doh2023r3337}. The accepted nutrition, ingredient and product-form evidence is then mapped to one of four product-level screening outcomes. \textit{Warning criteria met} indicates that the available and accepted evidence satisfies at least one encoded draft R3337 warning criterion. It does not indicate that a warning symbol was visually detected on the package. \textit{No warning criteria triggered} indicates that the available and accepted evidence is sufficient for screening but does not establish an encoded warning criterion and is not interpreted as confirmation of legal compliance. \textit{Review} indicates that relevant evidence is available but remains uncertain, ambiguous or conflicting, while \textit{Insufficient evidence} indicates that evidence required to apply the screening rules is unavailable. \textit{Review} and \textit{Insufficient evidence} represent abstention outcomes rather than forced automatic decisions. Qualifying unresolved evidence from the abstention outcomes proceeds to the selective VLM evidence-review stage described in Section~4.5. 

\subsection{Selective Vision-Language Model Evidence Review}

\begin{figure}[!htbp]
    \centering
    \includegraphics[width=\textwidth]{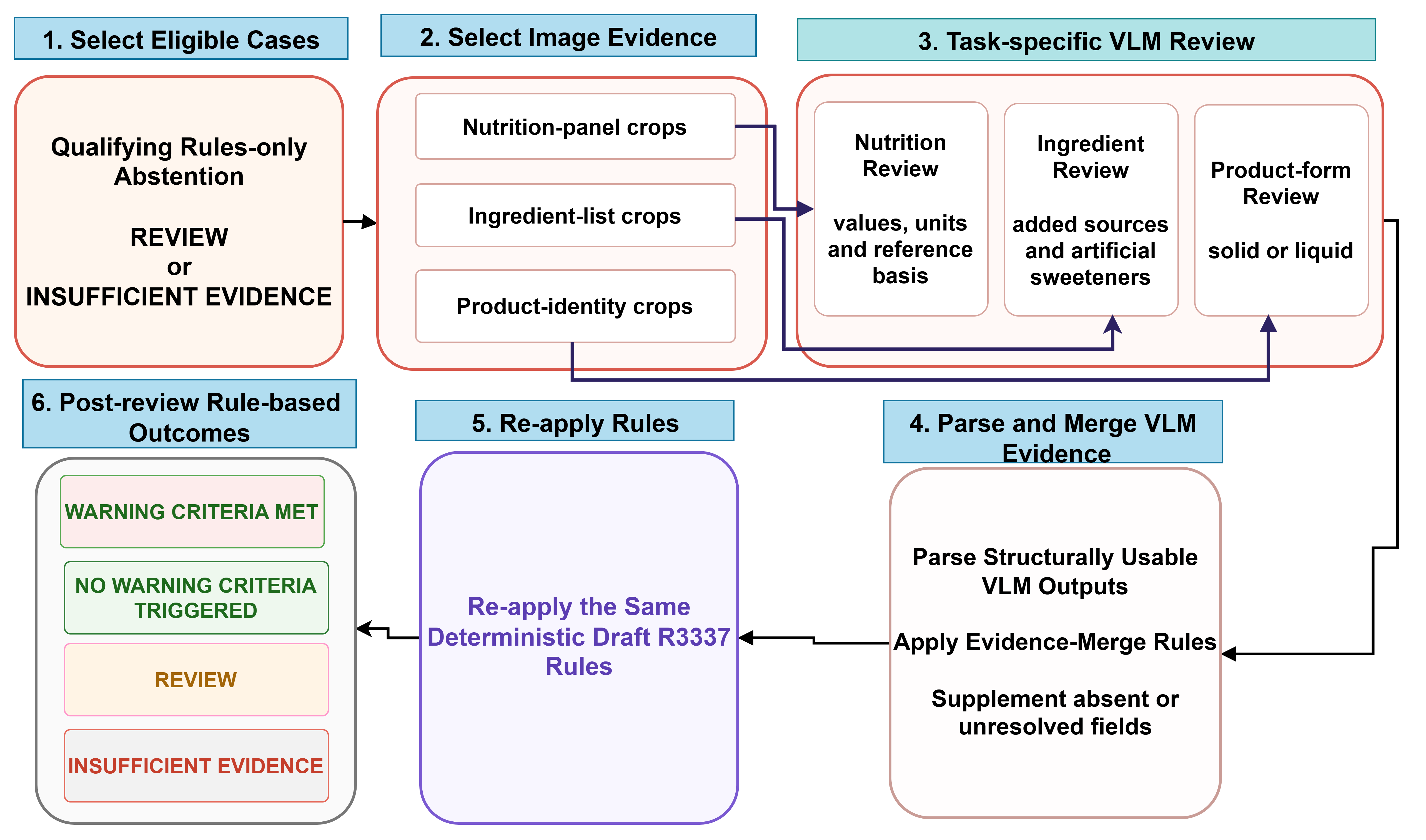}
    \caption{Selective VLM evidence-review and rule-reapplication workflow.}
    \label{fig:vlm_review_workflow}
\end{figure}

Existing automatic rules-only outcomes are not reconsidered. Figure~\ref{fig:vlm_review_workflow} presents the six-stage selective VLM evidence-review process for qualifying rules-only abstentions.

\begin{enumerate}
    \item \textbf{Select eligible cases.} Products assigned \textit{Review} or \textit{Insufficient evidence} proceed to VLM review only when they contain qualifying unresolved nutrition, ingredient or product-form evidence.

    \item \textbf{Select image evidence.} Relevant nutrition-panel, ingredient-list or product-identity crops are selected for the unresolved evidence. The VLM reassesses this evidence rather than making an independent product-level screening decision.

    \item \textbf{Task-specific VLM review.} The same Qwen3-VL 8B Instruct model is used through separate task-specific review routines for nutrition evidence, ingredient evidence and product form. The nutrition review examines unresolved nutrient values, units and reference bases. The ingredient review examines added sugar, added sodium, added saturated-fat sources and artificial sweeteners, while the product-form review assesses whether package evidence supports a solid or liquid classification.

    \item \textbf{Parse and merge VLM evidence.} Each task-specific review returns a structured machine-readable response. Malformed, incomplete or unresolved responses are rejected, and structurally usable responses are not assumed to be factually correct. Under the evidence-merge rules, accepted VLM evidence supplements only absent or unresolved fields, while existing accepted rules-only evidence is retained.

    \item \textbf{Reapply rules.} The merged evidence is reassessed using the same deterministic draft R3337 rules in Section~4.4. Existing evidence requirements are retained, so a VLM-recovered nutrient value cannot establish a sugar, sodium or saturated-fat warning without the corresponding qualifying ingredient evidence.

    \item \textbf{Post-review rule-based outcomes.} Rule reapplication produces the post-review product-level outcome. Unresolved cases continue to abstain rather than being forced into an automatic outcome. Post-review outcomes are retained separately from initial rules-only outcomes for evaluation.
\end{enumerate}

During development, Phi-3.5-Vision Instruct~\cite{abdin2024phi3} and Qwen3-VL 8B Instruct~\cite{bai2025qwen3vl} were examined. Neither model was trained or fine-tuned in this study. Both pretrained models were used for inference. Phi-3.5-Vision Instruct was used only during exploratory testing and was not retained after producing confident but incorrect associations between nutrition-table rows and values. Qwen3-VL 8B Instruct was retained for the final evidence-review experiments and deployed locally through Ollama. Manual reference outcomes are withheld during VLM review, evidence merging and deterministic rule reapplication and are used only during evaluation. The factual correctness of structurally usable VLM evidence is evaluated separately using the evidence-level measures in Section~4.6.

\subsection{Manual Validation and Evaluation Measures}

A 65-product manual validation set is constructed using stratified sampling across the four initial rules-only outcome categories. The set contains 22 randomly selected products assigned \textit{Warning criteria met}, nine products assigned \textit{No warning criteria triggered}, 17 randomly selected products assigned \textit{Review} and 17 randomly selected products assigned \textit{Insufficient evidence}. These rules-only outcomes are used only as sampling strata rather than as manual reference outcomes. The validation set is not intended to estimate the prevalence of the four outcome categories in the complete dataset. Manual reference review uses all available full-package photographs for each product and applies the same operational draft R3337 screening protocol described in Section~4.4. Reference outcomes are assigned from the complete visible package evidence.

Where a complete and readable ingredient declaration establishes that none of the qualifying ingredient conditions is present, manual review can assign \textit{No warning criteria triggered} even when the corresponding nutrient values are unavailable. This differs from the automated \textit{Not detected} ingredient result in Section~4.3, which indicates only that no configured qualifying term was recovered from OCR-derived text and is not treated as proof of absence.

Manual review assigns 37 products to \textit{Warning criteria met}, 23 to \textit{No warning criteria triggered} and five to \textit{Insufficient evidence}, resulting in 60 manually decidable products. The five unresolved products lack the evidence required for a definitive reference outcome after all available package images are examined. Product-specific evidence sources, reasons for unresolved outcomes and evidence-level reference fields for nutrition values, ingredient evidence and product form are recorded. All manual reference information is withheld from the rules-only workflow, selective VLM review and deterministic rule reapplication and is used only during evaluation.

Product-level evaluation is performed on the 60 manually decidable products using the following measures. To reflect uncertainty associated with the validation-set size, 95\% Wilson confidence intervals (CIs) are reported for coverage and selective accuracy.

\begin{itemize}
    \item \textbf{Coverage:} the proportion of products assigned an automatic outcome, namely \textit{Warning criteria met} or \textit{No warning criteria triggered}.
    \item \textbf{Selective accuracy:} the number of correct automatic outcomes divided by the total number of automatic outcomes, with abstentions excluded from the denominator.
    \item \textbf{Abstention rate:} the proportion of products assigned \textit{Review} or \textit{Insufficient evidence}.
    \item \textbf{False warning:} an automated \textit{Warning criteria met} outcome when the manual reference is \textit{No warning criteria triggered}.
    \item \textbf{Unsafe warning omission:} an automated \textit{No warning criteria triggered} outcome when the manual reference is \textit{Warning criteria met}.
\end{itemize}

Evidence-level VLM evaluation assesses usability and correctness for nutrition, ingredient and product-form evidence. Nutrition correctness requires agreement on the nutrient value, unit and reference basis, with strict exact matching also reported. Ingredient correctness is calculated among usable category decisions and product-form correctness among usable solid or liquid classifications.

\subsection{Computational Environment}

All computational experiments are conducted on the UWC High-Performance Computing Cluster. The GPU-enabled compute environment uses an AMD EPYC-Milan processor and an NVIDIA L4 GPU with approximately 23~GB of GPU memory. Evidence extraction, product-level aggregation and deterministic screening are implemented in Python 3.10. 

\section{Results and Evaluation}

\subsection{Region Detection and Evidence Availability}

On the validation and held-out test subsets, respectively, detector precision was 0.846 and 0.889, while recall was 0.774 and 0.952. Mean average precision (mAP) at an intersection-over-union (IoU) threshold of 0.50 was 0.810 and 0.947, while mAP averaged across IoU thresholds from 0.50 to 0.95 was 0.635 and 0.757. Validation and test results are reported separately, as validation performance informed checkpoint selection while the product-disjoint test subset was reserved for final evaluation. Full-dataset inference produced 4,552 crops from 3,106 images, comprising 1,976 product-identity, 1,383 ingredient-list and 1,193 nutrition-panel regions. At least one target region was detected for every product.

OCR processing was completed for all 4,552 detected crops, although not every crop contained usable recognised text. Processing the original and padded nutrition-panel renderings produced 2,386 crop-rendering records containing 82,682 recognised tokens. Tokens were recovered from 2,363 renderings, while 23 contained no recognised tokens. Structured nutrition evidence was recovered for 334 products. Accepted total-sugar evidence was available for 170 products, total-fat for 157, saturated-fat for 156, sodium for 158 and salt for four. Usable recognised ingredient text was available for 386 products. Fifty-four products had no detected ingredient-list region, while two had a detected ingredient region with no recognised text. Product-form assessment classified 222 products as \textit{Solid} and 51 as \textit{Liquid}, while 169 remained \textit{Review}.

\subsection{Rules-Only Product-Level Results}

Across all 442 products, the rules-only workflow produced 83 automatic outcomes, an operational decision rate of 18.8\%, and abstained on 359 products. These full-dataset results describe workflow behaviour rather than accuracy, as the complete dataset was not manually validated. Among the 60 manually decidable validation products, the workflow produced 31 automatic outcomes, of which 30 agreed with the manual reference. This corresponded to 51.7\% coverage (95\% CI 39.3--63.8\%) and 96.8\% selective accuracy (95\% CI 83.8--99.4\%), while missing, uncertain or conflicting required evidence contributed to a 48.3\% abstention rate. One false warning and no unsafe warning omissions were observed.

\subsection{Product-Level Results After Selective VLM Review}

Of the 359 products on which the rules-only workflow abstained, accepted Qwen3-VL evidence was incorporated for 232 products under the evidence-merge rules. Existing automatic rules-only outcomes remained unchanged. After evidence merging and rule reapplication, the product-level outcome changed for 144 products. Ninety-four abstentions became automatic outcomes, while 50 products moved between \textit{Review} and \textit{Insufficient evidence}. Across the dataset, the operational decision rate increased from 18.8\% to 40.0\%. These changes describe workflow behaviour rather than accuracy, as the dataset was not manually validated.

Among the 60 manually decidable validation products, selective VLM review resolved eight additional cases, increasing coverage to 65.0\% (95\% CI 52.4--75.8\%) and reducing the abstention rate from 48.3\% to 35.0\%. Overall, 38 of 39 automatic outcomes agreed with the manual reference, giving 97.4\% selective accuracy (95\% CI 86.8--99.5\%). The single false warning remained from the rules-only workflow, with no unsafe warning omissions after selective review. Tables~\ref{tab:full_dataset_outcomes} and~\ref{tab:manual_validation_results} summarise the full-dataset outcome changes and validation-set performance, respectively.

\begin{table}[!htbp]
\centering
\caption{Full-dataset outcomes before and after selective VLM evidence review.}
\label{tab:full_dataset_outcomes}
\small
\setlength{\tabcolsep}{4pt}
\setlength{\arrayrulewidth}{0.4pt}
\renewcommand{\arraystretch}{1}

\begin{tabular*}{\linewidth}{
    |>{\raggedright\arraybackslash}p{0.52\linewidth}
    @{\extracolsep{\fill}}
    r
    r|
}
\hline
\textbf{Outcome or measure} &
\textbf{Rules-only} &
\shortstack{\textbf{After selective}\\\textbf{VLM review}} \\
\hline
WARNING CRITERIA MET & 71 & 115 \\
NO WARNING CRITERIA TRIGGERED & 12 & 62 \\
REVIEW & 149 & 109 \\
INSUFFICIENT EVIDENCE & 210 & 156 \\
\hline
Automatic outcomes & 83 (18.8\%) & 177 (40.0\%) \\
Abstentions & 359 (81.2\%) & 265 (60.0\%) \\
\hline
\end{tabular*}
\end{table}

\begin{table}[!htbp]
\centering
\caption{Performance on the 60 manually decidable validation products before and after selective VLM review.}
\label{tab:manual_validation_results}
\small
\setlength{\tabcolsep}{4pt}
\setlength{\arrayrulewidth}{0.4pt}
\renewcommand{\arraystretch}{1}

\begin{tabular*}{\linewidth}{
    |>{\raggedright\arraybackslash}p{0.42\linewidth}
    @{\extracolsep{\fill}}
    r
    r|
}
\hline
\textbf{Measure} &
\textbf{Rules-only} &
\shortstack{\textbf{After selective}\\\textbf{VLM review}} \\
\hline
Coverage & 31/60 (51.7\%) & 39/60 (65.0\%) \\
Selective accuracy & 30/31 (96.8\%) & 38/39 (97.4\%) \\
Abstention rate & 29/60 (48.3\%) & 21/60 (35.0\%) \\
False warnings & 1 & 1 \\
Unsafe warning omissions & 0 & 0 \\
Newly resolved products & -- & 8 \\
\hline
\end{tabular*}
\end{table}

\FloatBarrier

\subsection{Evidence-Level Reliability of VLM-Recovered Evidence}

Tables~\ref{tab:qwen_nutrition_results} and~\ref{tab:qwen_ingredient_form_results} report the usability and correctness of Qwen3-VL 8B Instruct evidence, showing that structurally usable outputs were not always factually correct.

\begin{table}[!htbp]
\centering
\caption{Nutrition-evidence reliability of Qwen3-VL 8B Instruct.}
\label{tab:qwen_nutrition_results}
\small
\setlength{\tabcolsep}{4pt}
\setlength{\arrayrulewidth}{0.4pt}
\renewcommand{\arraystretch}{1}

\begin{tabular*}{\linewidth}{
    |>{\raggedright\arraybackslash}p{0.34\linewidth}
    @{\extracolsep{\fill}}
    r
    r
    r|
}
\hline
\textbf{Evaluation} &
\textbf{Usable/evaluated} &
\textbf{Correct} &
\textbf{Correctness} \\
\hline
Nutrition overall & 61/76 & 47/61 & 77.0\% \\
Strict exact match & 61/76 & 46/61 & 75.4\% \\
Total sugar & 21 & 14/21 & 66.7\% \\
Saturated fat & 20 & 15/20 & 75.0\% \\
Sodium & 20 & 18/20 & 90.0\% \\
\hline
\end{tabular*}
\end{table}

\begin{table}[!htbp]
\centering
\caption{Ingredient and product-form evidence reliability of Qwen3-VL 8B Instruct.}
\label{tab:qwen_ingredient_form_results}
\small
\setlength{\tabcolsep}{4pt}
\setlength{\arrayrulewidth}{0.4pt}
\renewcommand{\arraystretch}{1}

\begin{tabular*}{\linewidth}{
    |>{\raggedright\arraybackslash}p{0.34\linewidth}
    @{\extracolsep{\fill}}
    r
    r
    r|
}
\hline
\textbf{Evidence category} &
\textbf{Usable/decided} &
\textbf{Correct} &
\textbf{Correctness} \\
\hline
Ingredient evidence overall & 75/108 & 66/75 & 88.0\% \\
Added sugar & 21 & 19/21 & 90.5\% \\
Added sodium & 21 & 17/21 & 81.0\% \\
Added saturated fat & 16 & 13/16 & 81.2\% \\
Artificial sweetener & 17 & 17/17 & 100.0\% \\
\hline
Product form & 20/33 & 19/20 & 95.0\% \\
\hline
\end{tabular*}
\end{table}


\subsection{Case Analysis}

Selected validation products illustrate how VLM-recovered evidence affected product-level outcomes and are not intended to represent the full validation set or complete dataset. For SaveMor Coconut Biscuits, the VLM recovered total sugar as 17.8~g per 100~g, matching the manual reference value, while qualifying added-sugar evidence supported a \textit{Warning criteria met} outcome through the sugar criterion. SPAR Split Peas illustrates that a correct product-level outcome can still contain incorrect recovered evidence. The VLM recovered total sugar as 0.7~g per 100~g, while manual review recorded 2.7~g per 100~g. Both values were below the applicable solid-product threshold, so the error did not alter the final \textit{No warning criteria triggered} outcome. A similar sugar-extraction error could affect the product-level outcome if the correct value exceeded the applicable threshold and qualifying added-sugar evidence was present. These cases show why product-level performance should be interpreted alongside evidence-level correctness.

\section{Discussion}

\subsection{Evidence Recovery Limited Automatic Coverage}

Despite strong held-out detector performance, automatic coverage was limited by incomplete nutrition, ingredient and product-form evidence. These limitations are consistent with the packaging-specific OCR challenges reported by Nagayi et al.~\cite{nagayi2026ocr} and the broader image-condition challenges identified by Guimarães et al.~\cite{guimaraes2023review}. Manual review also encountered blurred sections and visible skew, although their effect on individual extraction errors was not isolated. Region localisation restricted OCR to detected package areas, following the detection-before-recognition approach used by Saputra et al.~\cite{saputra2024bounding}. Structured screening also required nutrient labels to be associated with the correct values, units and reference bases, reflecting the importance of spatial association demonstrated by Fuentes et al.~\cite{fuentes2022transcription}, while ingredient and product-form uncertainty introduced further abstentions. These findings support retaining unresolved cases rather than treating missing or uncertain evidence as a negative result.

\subsection{Coverage, Selective Accuracy and Abstention}

Selective VLM review resolved eight additional validation products without introducing a new false warning or unsafe warning omission. The small number of resolved cases, however, limits broader conclusions about this improvement. Geifman and El-Yaniv~\cite{geifman2019selectivenet} examined coverage and error risk in selective prediction, while Whitehead et al.~\cite{whitehead2022reliable} showed the value of abstaining when confidence is insufficient in visual question answering. In this workflow, abstention is determined by deterministic evidence requirements rather than a probabilistic confidence score, so confidence calibration was not evaluated. The rules-only and post-review results provide two operating points for the coverage-abstention trade-off, but not a full risk-coverage curve. Coverage should therefore be interpreted together with selective accuracy and error type.

\subsection{Benefits and Risks of Vision-Language Evidence Review}

Selective VLM review recovered useful evidence for some unresolved cases, but reliability varied across nutrition, ingredient and product-form tasks. Nutrition evidence showed the lowest correctness of the three evaluated evidence types, indicating that exact nutrient recovery remained more difficult than the broader ingredient and product-form decisions in this evaluation. The case analysis also showed that an incorrect recovered nutrient value could coexist with a correct product-level outcome when the error did not change the applicable warning criterion. This distinction is important, as high selective accuracy at product level does not establish that all supporting evidence is correct. Liu et al.~\cite{liu2024ocrbench} similarly reported useful multimodal text-recognition capabilities alongside continuing weaknesses on challenging text-related image tasks, while Assiri et al.~\cite{assiri2025nutrition} demonstrated the applicability of multimodal models to nutrition extraction from food-label images. The causes of individual VLM errors were not isolated in this study, so the observed errors cannot be attributed to a single failure source. The evidence-level results therefore support retaining the VLM as a selective evidence reviewer rather than an independent product-level decision-maker.

\subsection{Relationship to Manual Draft R3337 Assessment} 

Unlike the draft R3337 assessments of Abdool Karim et al.~\cite{abdoolkarim2025compliance} and Mamabolo et al.~\cite{mamabolo2026warning}, which used captured or manually coded product information, this study addresses evidence recovery directly from food-package photographs. The workflow therefore supports screening rather than automated regulatory assessment, and a \textit{No warning criteria triggered} outcome does not establish legal compliance.

\section{Limitations}

The stratified 65-product validation set was not designed to estimate outcome prevalence, and five products remained unresolved after all package images had been reviewed. The same set supported product-level and evidence-level evaluation, while post-review outcomes for the 442-product dataset were not manually verified. Structurally usable VLM outputs could contain incorrect evidence. Broader evaluation using an independent dataset and systematic error-sensitivity analysis is therefore needed to assess robustness and generalisation. Evidence availability differed across nutrition, ingredient and product-form information, with limitations in nutrient-row association, ingredient recovery and unclear preparation or product-form evidence. Salt evidence was particularly scarce. The study used one South African food-package image collection, so generalisation to other products, retailers, languages and image conditions remains untested. The study did not include a full-image OCR baseline and therefore does not isolate the effect of region localisation. R3337 remained in draft form at the time of the study, and its criteria may change before the regulations are finalised. The final evidence-review evaluation used one retained VLM, so the findings do not establish generalisability across different VLMs. The dataset cannot be publicly released as it forms part of an ongoing project, while the code is available on request from the corresponding author.

\section{Conclusion}

This study developed and evaluated a region-guided, abstention-aware workflow for draft R3337 screening from South African food-package photographs. On the 60 manually decidable validation products, rules-only screening achieved 51.7\% coverage and 96.8\% selective accuracy, while selective Qwen3-VL 8B Instruct review increased coverage to 65.0\% and selective accuracy to 97.4\%, with one false warning and no unsafe warning omissions. Evidence-level evaluation showed that structurally usable VLM outputs could still contain incorrect supporting evidence. Selective VLM review can therefore supplement unresolved evidence, while deterministic rule reapplication and explicit abstention remain important. Further work should evaluate cross-checking between OCR-derived and VLM-recovered evidence, agreement across multiple VLMs, and improved nutrition-row association, ingredient recovery and product-form determination. The workflow provides decision support rather than final legal or regulatory determinations.

\bibliographystyle{splncs04}
\bibliography{references}

\end{document}